\documentclass[letterpaper, 10 pt, conference]{ieeeconf}  %

\IEEEoverridecommandlockouts                              %

\usepackage{times}
\usepackage{epsfig}
\usepackage{graphicx}
\usepackage{amsmath}
\usepackage{amssymb}

\usepackage[colorlinks,
    breaklinks=true,
    citecolor=blue]{hyperref}

\hypersetup{
    pdftoolbar=true,        %
    pdfmenubar=true,        %
    pdffitwindow=false,     %
    pdfstartview={FitH},    %
    pdflang = {en-US},
    pdftitle={High-Degrees-of-Freedom Dynamic Neural Fields for Robot Self-Modeling and Motion Planning},    %
    pdfauthor={Lennart Schulze, Hod Lipson},     %
    pdfsubject={High-Degrees-of-Freedom Dynamic Neural Fields for Robot Self-Modeling and Motion Planning},   %
    pdfkeywords={neural fields, motion planning, computer vision, machine learning, robotics}, %
    pdfnewwindow=true,      %
    pdfstartpage=1,
    pdfdirection=L2R,
    pdfdisplaydoctitle=true,
    pdfpagelabels=true,
    bookmarks=true,
    bookmarksnumbered=true,
    bookmarksopen=false,
}

\title{\LARGE \bf
High-Degrees-of-Freedom Dynamic Neural Fields \\for Robot Self-Modeling and Motion Planning
}

\author{Lennart Schulze$^{1}$ and Hod Lipson$^{2}$%
\thanks{$^{1}$Department of Computer Science, Columbia University, New York, NY 10027, USA.
        {\tt\small lennart.schulze@columbia.edu}}%
\thanks{$^{2}$Department of Mechanical Engineering, Columbia University, New York, NY 10027, USA.
        {\tt\small hod.lipson@columbia.edu}}%
}

\usepackage[backend=bibtex,
    sorting=none,
    citestyle=numeric,
    bibstyle=ieee,
    maxbibnames=9,
    doi=false,isbn=false,url=true,eprint=false]{biblatex}
\begin{document}

\maketitle
\thispagestyle{empty}
\pagestyle{empty}

\begin{abstract}

A robot self-model is a task-agnostic representation of the robot's physical morphology that can be used for motion planning tasks in the absence of a classical geometric kinematic model. In particular, when the latter is hard to engineer or the robot's kinematics change unexpectedly, human-free self-modeling is a necessary feature of truly autonomous agents. In this work, we leverage neural fields to allow a robot to self-model its kinematics as a neural-implicit query model learned only from 2D images annotated with camera poses and configurations. This enables significantly greater applicability than existing approaches which have been dependent on depth images or geometry knowledge. To this end, alongside a curricular data sampling strategy, we propose a new encoder-based neural density field architecture for dynamic object-centric scenes conditioned on high numbers of degrees of freedom (DOFs). In a 7-DOF robot test setup, the learned self-model achieves a Chamfer-L2 distance of 2\% of the robot's workspace dimension. We demonstrate the capabilities of this model on motion planning tasks as an exemplary 
downstream application.

\end{abstract}

\section{Introduction}

Neural fields paired with differentiable rendering allow learning accurate 3D scene information from pose-annotated 2D images. This is achieved by overfitting a neural network to the scene observed from multiple camera views using a photometric reconstruction loss \cite{mildenhall2021nerf}. After training, the model can be used to render realistic images of the scene from novel camera views.
Due to the importance of scene representations in robotics, neural field extensions have evolved focusing on use cases in this area. While most of these approaches \cite{lee2022reconstruction, adamkiewicz2022vision, moreau2022lens, maggio2023loc} use neural fields to capture and utilize information about the robot's environment, such as for reconstruction, navigation, or localization tasks, here we propose to learn neural fields to represent - and control - the robot.

\begin{figure}[tpb]
      \centering
      \framebox{
      \includegraphics[scale=0.32]{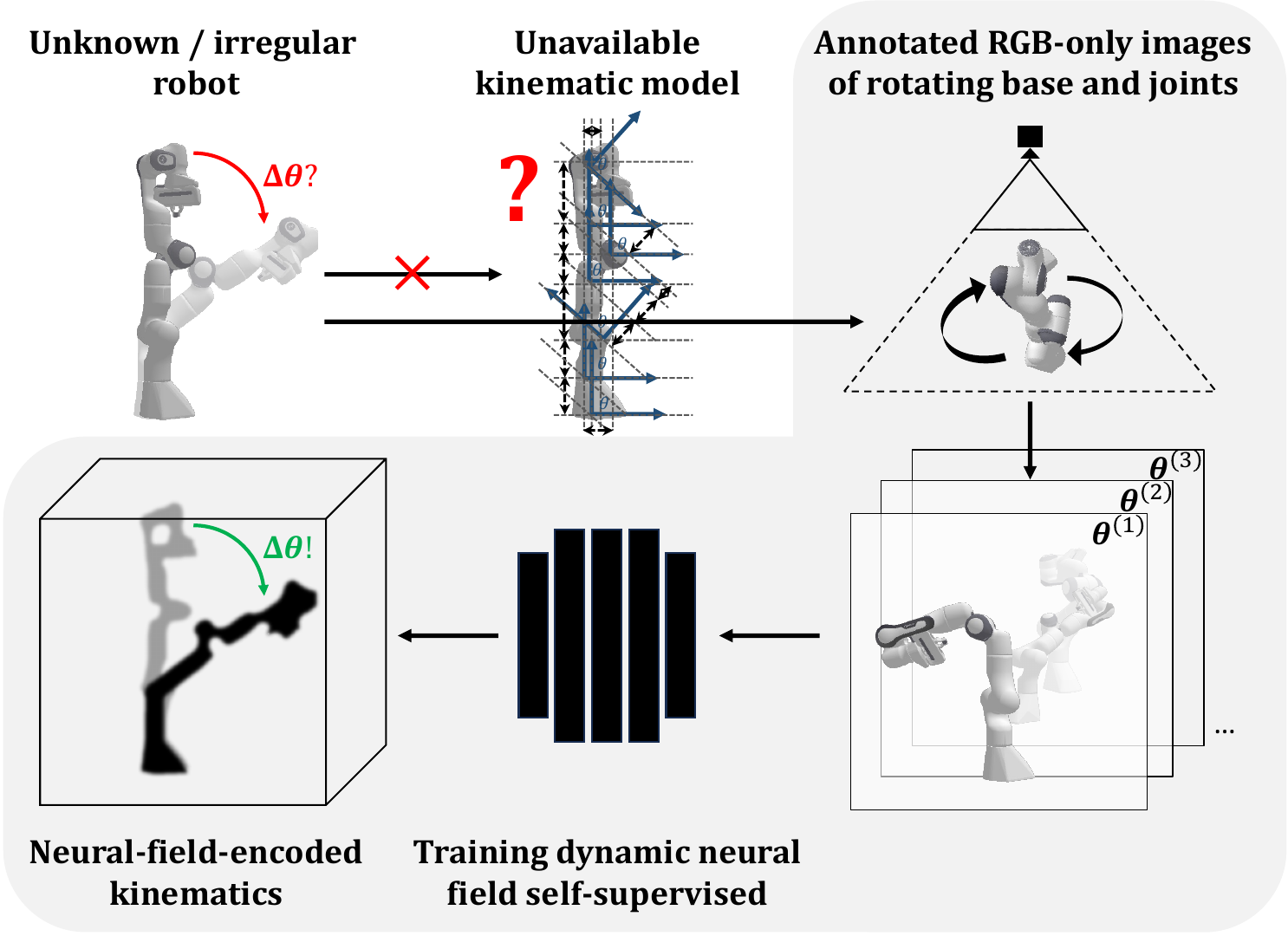}
      }
      \caption{\textbf{Overview of contributions (shaded):} When a kinematic model is unavailable for the robot, 1) our method to collect curricular annotated depth-free image data can be used instead to train 2) a high-DOFs dynamic neural density field which the robot uses as a self-model. 3) Its forward and inverse kinematics capabilities enable motion planning applications.}
      \label{fig:overview}
\end{figure}

We solve the task of robot self-modeling, the (robot's) ability to acquire a representation of the robot's kinematics from observing its behavior without human interference%
. Similar to a mental image of oneself, self-models can continually be updated to reflect the state of the robot. %
This renders them advantageous over classical geometric kinematic models, which are
usually engineered once, may be mismatched to the current state of the robot, and are unavailable for unknown robots \cite{stone1987kinematic}.  
For these reasons, learning-based approaches to robot self-modeling emerged. Despite functional, a major drawback is their dependence on supervised samples or, in the self-supervised case, depth annotations in the training distribution.
These requirements hinder the readiness of target applications of self-modeling in real-world scenarios where this information is not available, such as after damage to the robot's body during deployment. 
In particular, a recent approach \cite{chen2022fully} to learn a full-body kinematic forward model as a neural-implicit representation requires images annotated with depth values from RGB-D cameras.
In this work, we propose to solve this obstacle by learning neural fields in a self-supervised manner directly from 2D images only annotated with camera parameters and the dynamic configuration.
Consequently, we approach the task of learning a neural-implicit full-body kinematic model from unlabeled kinematic data %
through training a dynamic neural field that offers downstream compatibility (Fig. \ref{fig:overview}).

We achieve this by introducing a new type of \textit{dynamic} neural field. Previous work \cite{pumarola2021dnerf,park2021nerfies,tretschk2021non} has extended the static-scene setup of neural radiance fields \cite{mildenhall2021nerf} by establishing time as an additional input dimension next to 3D coordinates, which together are mapped to density and color values. %
In contrast, in this work we introduce a high number of degrees of freedom (DOFs) that in complex interdependence change local parts of the scene as the conditioning variables for a coordinate-to-density map, which has not been done for %
robotic applications.
Different from methods using deformation from a canonical representation \cite{peng2021animatable}, we propose a DOF-encoder-based dynamic neural density field, which is suitable for modeling %
complex changing scenes beyond robotics.

In summary, this work contributes the following:
\begin{itemize}
    \item We introduce a curricular data sampling method and neural network architecture %
    to represent high-DOFs object-centric scenes as dynamic neural density fields.
    \item %
    We use our method to visually learn the first robot self-model without depth information and from a single camera view, and quantify its quality experimentally.
    \item Extending \cite{chen2022fully}, we discuss and demonstrate downstream applications of neural-field self-modeled kinematics in motion planning.
\end{itemize}

\section{Background and Related Work}

\textbf{Robot self-modeling.}
A self-model is a task-agnostic, general-purpose representation of a robot's physical shape and structure that can be acquired at any time and continually updated %
without a human in the loop \cite{kwiatkowski2022deep, dearden2008developmental}. 
The objective of enabling machines to produce a cognitive model of themselves to guide their behavior has been inspired by similar behavior in human beings \cite{rochat2003five}. In practice, whenever a geometric kinematic model%
, which captures the spatial relations and physical constraints of the robot's links and joints manually as a result of simulation and engineering, is unavailable, the ability to self-model is required. In particular, when the robot's kinematics are altered, for instance through damage or undocumented body manipulation, the robot can learn an updated self-model without the need to manually re-devise the kinematic model
\cite{bongard2006resilient, bongard2004automated}.

Approaches to robot self-modeling have leveraged analytical, probabilistic, and evolutionary methods \cite{bongard2006resilient, gold2009prob, bongard2004automated}. Learning-based approaches to implicitly represent self-models were first presented in \cite{kwiatkowski2019task}, necessitating training samples that are labeled with the end effector position.
Similarly, certain approaches \cite{hang2021manipulation, jiang2017new} pre-determine the set of parameters to learn for a system, identified based on prior knowledge about the shape or function. 
The most recent, partially self-supervised approach, which constructs an agnostic self-model without such information \cite{chen2022fully}, still requires depth information, which is used to learn an SDF-based occupancy query model. 
In all approaches, data acquisition plays a crucial role, with strategies ranging from entirely random \cite{chen2022fully}, to interactive \cite{bohg2017interactive}, and targeted-exploratory \cite{hang2021manipulation, amos2018learning}. 
This work builds on the agnostic, neural-implicit class of representation proposed in \cite{chen2022fully} and removes the depth requirement using neural fields, while introducing a curricular-random training data acquisition strategy.

\textbf{Neural (radiance) fields.}
A neural field is a continuous map from any spatial coordinate in 3D space $\mathbf{x}=\begin{bmatrix}x,y,z\end{bmatrix}^T$ to a scalar or vector. 
In neural \textit{radiance} fields (NeRF) \cite{mildenhall2021nerf}, each point is assigned a tuple of density and color. The map is parameterized via a neural network $\Phi$, such as a multi-layer perceptron (MLP), overfit to the specific scene,
\begin{equation}
 f_{\Phi} : (\mathbf{x},\mathbf{d}) \to (\sigma, \mathbf{c})
\end{equation}
where $\mathbf{x}\in \mathbb{R}^3$ is the coordinate vector, %
$\mathbf{d}\in[0,1]^3,\|\mathbf{d}\|=1$ 
is the unit viewing direction,  $\sigma \in [0,\infty)$ is the predicted density, and $\mathbf{c} \in [0,1]^3$ is the predicted RGB color. Both to march a ray through the scene to obtain point coordinates $\mathbf{x}$ and to compute $\mathbf{d}$, the camera pose $\boldsymbol{^wT_c}$ is used.

A NeRF is a neural-implicit representation of a scene since novel views can be rendered by querying the learned map, without the need to store 3D information explicitly, such as in point clouds or voxels.
From the field over the 3D space, 2D projections to images from arbitrary camera views are rendered via volume rendering \cite{kajiya1984ray, max1995optical}: The color of a pixel $\mathbf{C}$ is computed by integrating the product of color, density, and visibility of the points residing on the ray $\mathbf{r}$ that was marched through the scene from the projection plane within the depth view bounds. The visibility $T_i$ of a point depends on the density values of the points between the projection plane and that point.
Using quadrature, the integral is approximated on $N$ points, which are sampled in a stratified manner from bins on the ray, as follows:
\begin{equation}
    \hat{\mathbf{C}}(\mathbf{r}) = \sum_{i=1}^N 
    \hat{T}_i ~ \alpha(\sigma_{\Phi}(\mathbf{x^{(i)}})\delta_i) ~
    \mathbf{c}_{\Phi}(\mathbf{x^{(i)}},\mathbf{d}) \\
\end{equation}
\begin{equation}
    \hat{T}_i = \exp  \left ( -\sum_{j=1}^{i-1} \sigma_{\Phi}(\mathbf{x^{(j)}})\delta_j \right )
\end{equation}
Here, $\mathbf{r}=\mathbf{o}+t\mathbf{d}$ is the pixel-corresponding ray marched through the scene scaled by depth $t$; $\alpha(\sigma)=1-\exp(-\sigma)$ maps density values into the range $[0,1]$; and $\delta_i=t_{i+1}-t_i$ is the distance between adjacent points %
on the ray. 
The differentiable nature of volume rendering allows the MLP to be trained by minimizing a photometric reconstruction loss between training images and renders from the same poses. %

Dynamic neural fields have emerged to represent scenes with changing components, for instance over a single time dimension \cite{pumarola2021dnerf,park2021nerfies,tretschk2021non,li2021sceneflow,xian2021space,gao2021dynamic,noguchi2021narf,li2022neural}. In extension, numerous works have aimed at modeling controllable human bodies, customarily using prior knowledge or annotations about their shape or multi-view video training data \cite{correia20233d,sun2022human,peng2021animatable,liu2021neuralactor,weng2022humannerf,li2022tava}. Related to our work, we propose a new dynamic neural field architecture geared towards shape-unknown objects with many DOFs that are interdependent, trained on single-view images only.

\section{Method}

\begin{figure*}
\begin{center}
\vspace{2mm}
\includegraphics[scale=0.49]{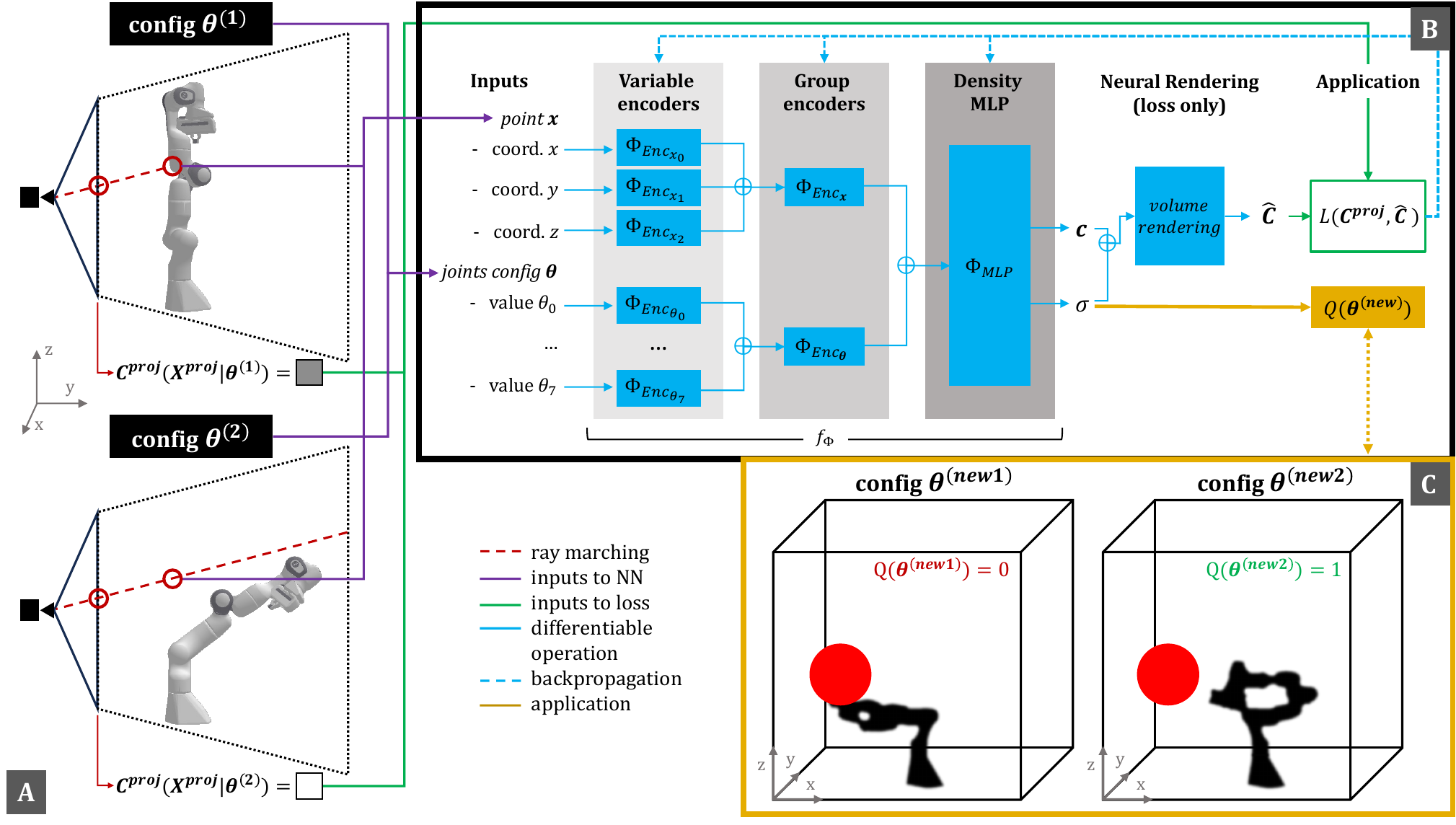}
\end{center}
   \caption{\textbf{Overview of proposed method:} A) Training images of robot in different configurations: While the point coordinates and annotated configuration are the inputs to the neural network, the true color of the pixel is used for the reconstruction loss. B) Neural network architecture: The DOF values and spatial coordinates are individually encoded, concatenated and group-wise encoded, and concatenated and processed to an output density. %
   C) The trained neural density field is used to evaluate the validity of a configuration relative to an obstacle by predicting the densities of points queried from its volume.}
\label{fig:nn}
\end{figure*}

\subsection{High-DOFs Dynamic Neural Density Field}

We propose to extend neural fields to dynamic scenes in which changes are anchored in interdependent DOFs of the object in the scene.
For this purpose, we condition the map $f_\Phi$ on the $k$-dimensional configuration of the object $\boldsymbol{\theta}=\begin{bmatrix}\theta_0, \ldots, \theta_{k-1}\end{bmatrix}^T\in\mathbb{R}^k$ that causes the observed changes:
\begin{equation}
    f_{\Phi} : (\mathbf{x},\boldsymbol{\theta}) \to (\sigma)
\end{equation}
To model the shape, the field does not assign color and is thus independent of the viewing direction. Nonetheless, color can be included for the purpose of training the model via a photometric loss.

Specifically, to learn a self-model of a robot, the map is conditioned on the joint configuration of the robot composed of $k$ joint values %
and thus learned as a $3+k$-dimensional neural field. We can subsequently compute density fields in novel configurations and render these from novel views. %

\textbf{Encoder-based architecture.}
We parameterize the map via a neural network based on \cite{mildenhall2021nerf},
that is an MLP with ReLU activations. Extending this architecture by adding $\boldsymbol{\theta}$ as input parameters produces unsatisfactory results. Consequently, we extend \cite{chen2022fully}'s approach and introduce separate encoders for the spatial and  conditioning input variables. This leverages the independence of the coordinates and the DOFs configuration and promises to learn useful representations resulting from the combinations of the constituents of each group.

Applying the shuffled curriculum learning approach introduced below, however, results in continual forgetting of previously learned relationships of DOFs as the training progresses, reducing performance on previously well reconstructed samples. For this reason, we introduce DOF-individual encoders, MLPs that encode each input variable. Since gradients flowing back to the weights of these MLPs will be zero when the joint values are zero in batches in which exclusively other DOFs are sampled, we argue this improves the memorability of useful features for the behavior introduced by each DOF individually. The outputs of the individual encoders are concatenated and chained to the group-based encoders. The complete model is described by:
\begin{multline}
    f_\Phi(\mathbf{x}, \boldsymbol{\theta}) = \\ \Phi_{\text{MLP}}\left (
    \Phi_{\text{Enc}_\textbf{x}}
    ( \bigoplus_{l=0}^{2} \Phi_{\text{Enc}_{x_l}}(x_l)  ), 
    \Phi_{\text{Enc}_{\boldsymbol{\theta}}}
     ( \bigoplus_{l=0}^{k-1} \Phi_{\text{Enc}_{\theta_l}}(\theta_l)  )
    \right )
\end{multline}
where $\oplus$ is the concatenation operator and $\Phi(x)=h_n \circ \ldots \circ h_1(x)$ is an $n$-layer MLP with ReLU activations. The architecture is shown in Fig. \ref{fig:nn}. 

Following \cite{mildenhall2021nerf}, we train two models of this type to separately model spatially coarse and fine predictions. The second model evaluates points on the ray sampled from regions of $t$ where the first model has resulted in higher density predictions. All points are used to render %
the projection of a ray. 
We find the sinusoidal positional encoding used in \cite{mildenhall2021nerf} to hinder the learning of the physical movement associated with traversing the DOF value ranges. Consequently, we substitute it with the $[-1,1]$-normalized
original joint values, resulting in $3+k$-dimensional inputs %
to the model.

\textbf{Learning from a one-camera setup.}
To circumvent the need for multiple cameras observing the robot to produce the neural field, which limits real-world applications, we harness the mobility of the robot's base. Given the robot's configuration $\boldsymbol{\theta}$ and the camera pose as camera-to-world transform $\boldsymbol{^wT_c}$, we enforce the first DOF to be the base rotation. For a front-facing camera pointing at the center of the robot with zero pitch and roll, rotating the object at its base is equivalent to rotating the camera about the upward axis. Hence, multi-view consistency for the density predictions can be enabled by assigning $(\boldsymbol{^wT_c})' \leftarrow \boldsymbol{R_z}(\theta_0)\boldsymbol{^wT_c}$ and $\theta_0' \leftarrow 0$.

\textbf{Curricular training data.}
Due to the large space of configurations and the serial dependence among the $k$ DOFs, the most distal joint's position depending on all $k-1$ previous DOFs, learning a high-DOFs neural field is difficult. Thus, the training data generation approach is crucial to the success of the model inferring the correct marginal influence of each DOF.
We propose a curriculum-learning-inspired sampling approach. For the set of all DOF indices $\Theta=\{l\}_{l=1}^{k-1}$, we compute the powerset containing all subsets of $\Theta$ and sort it in ascending order by magnitude, excluding the empty set: $S_\Theta = \{s\}_{s\subseteq\Theta}$. For each set of DOF indices $s\in S_\Theta$, samples are generated by uniformly randomly sampling values from the permissible ranges of the DOFs in $s$. 
The values of the remaining DOFs are fixed to zero.
Thus, we encourage learning the contribution of each DOF first by itself and then in combination with other DOFs in order of increasing complexity, until all DOFs are interacting. 
For example,
\begin{equation}
    \boldsymbol{\theta^{(1,4)}} = \begin{bmatrix}
    0 & \theta_1 \sim D_1 & 0 & 0 & \theta_4 \sim D_4 & \ldots & 0
    \end{bmatrix},
\end{equation}
where $D_i = \mathcal{U}(\theta^{(min)}_i,\theta^{(max)}_i)$.
We find shuffling the images such that images with different numbers of active DOFs lie in the same batch to improve the training performance. 

\textbf{Training.} 
We optimize the model via a photometric mean squared error (MSE) loss between ground-truth $\mathbf{C^{proj}}$ and rendered pixels $\mathbf{\hat{C}}$. Our experiments suggest that keeping the RGB output %
improves training performance. %
Nonetheless, the density prediction is the only output kept to be used in the self-model after training. To train density-output-only high-DOFs neural fields, the MSE loss may be used between binarized images and renderings with $\mathbf{c}$ set to black.

\begin{table*}[!htb]
\vspace{2mm}
\caption{Spatial distances between predicted and ground-truth self-model meshes in random test configurations and across test set.
}
\label{tab:results}
\begin{center}

\begin{tabular}{|r||ccccc|c|}
\hline
Distance metric & config a & config b & config c & config d & config e & test set (n=30) \\
\hline\hline
Chamfer-L2 (m) $\downarrow$ & .017 & .019 & .053 & .013 & .013 & .024 \\
Chamfer-L2 (\% of workspace-$z$) $\downarrow$ & 1.35 & 1.51 & 4.22 & 1.06 & 1.07 & 1.94 \\
Surface area IoU $\uparrow$ & .501 & .479 & .408 & .571 & .572 & .496 \\
Hull volume IoU $\uparrow$ & .685 & .607 &  .336 & .714 & .690 & .573 \\
\hline
\end{tabular}
\end{center}
\end{table*}

\begin{figure*}[!htb]
\begin{center}
\includegraphics[scale=0.48]{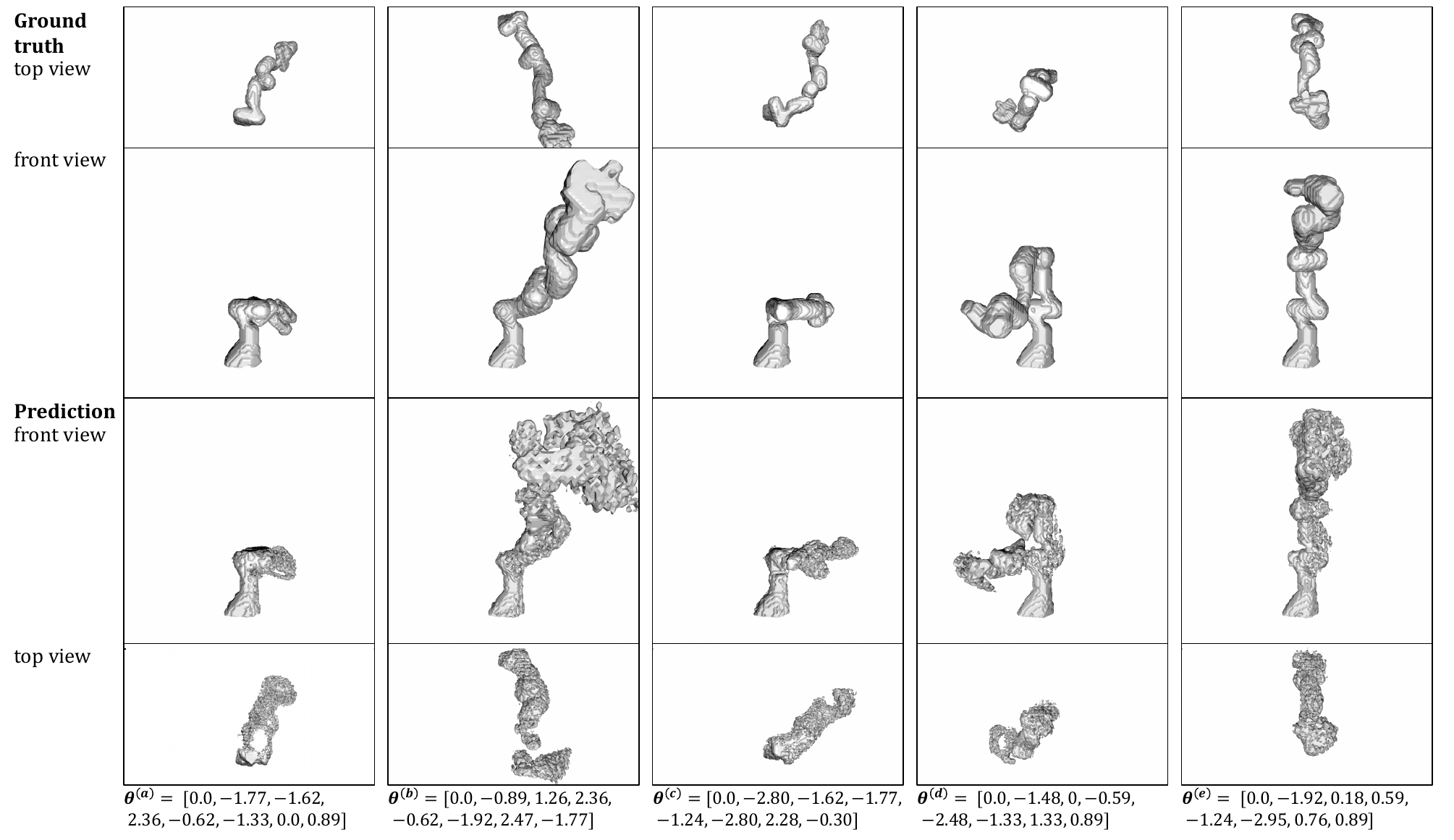}
\end{center}
   \caption{\textbf{Self-model results:} Predicted vs. ground-truth meshes, smoothed and reconstructed via marching cubes from point clouds generated by querying the high-DOFs neural density field in the given random test configuration. The configurations are shown in radians. Please also see suppl. video \cite{video}.}
\label{fig:meshes}
\end{figure*}

\subsection{Neural-Field Self-Model and Applications}
\textbf{Self-model.}
The trained map $f_\Phi$ is an implicit full-body kinematic model of the robot since it enables the reconstruction of its shape conditioned on its joint configuration. Learning this model only from annotated 2D images replaces the need to know the robot geometry altogether.

\textbf{Motion planning: Reaching a target via inverse kinematics.}
Extending \cite{chen2022fully}, we demonstrate motion planning as an inherent downstream application of the model. Due to the differentiable forward prediction of the density of a point given the configuration, we can compute the inverse kinematics, that is the configuration such that a point is occupied. 
For this purpose, the MLP parameters are fixed, and the input joint values are optimized via projected gradient descent (PGD) to minimize the delta to the desired density.

By choosing appropriate points, the robot can, for example, compute how to reach an object.
Furthermore, by selecting the initial configuration of the optimization to be the current configuration of the robot and acting in an obstacle-free environment, the inverse kinematics optimization steps can be cast as a path to reach the target. 
Precisely, given information about the target, we uniformly sample $N$ points from its surface $O_s=\{\mathbf{x^{(i)}}\}_{i=1}^N$ and query the robot's density on them,
leveraging that density on the surface above a threshold $\tau$ indicates touch.
In the fine model, starting from a no-touch configuration $\boldsymbol{\theta^{(0)}}$, we minimize the following loss, which will be $\leq0$ when the target is reached: 
\begin{equation}
    L(\boldsymbol{\theta}, O_s) = \min_{\textbf{x}\in O_s}  \left [-\alpha(f_\Phi(\mathbf{x},\boldsymbol{\theta}))  \right ] + \tau
\end{equation}
The final ReLU activation at $\Phi_{\text{MLP}}$'s output unit for $\sigma$ is removed to produce non-zero gradients. 
To enforce that the final joint configuration and every step of the optimization are within the joint limits, after each step of size $\eta$ the joint values are projected back into the $k$-dimensional ball representing the permissible ranges: 
\begin{equation}
\boldsymbol{\theta^{(j+1)}} = \Pi_{\boldsymbol{\theta^{(min)}}}^{\boldsymbol{\theta^{(max)}}} \left [ \boldsymbol{\theta^{(j)}} - \eta \frac{\partial L(\boldsymbol{\theta^{(j)}},O_s)}{\partial \boldsymbol{\theta^{(j)}}} \right ] 
\end{equation}
If only the inverse kinematics task is required, random initializations of the configuration can accelerate the optimization.%

\textbf{Motion planning: Configuration space.}
For more complex constraints and in the presence of obstacles, customary motion planning algorithms can be used with the self-model.
Any planning algorithm using the configuration space,
that is the binary map over the $k$-dimensional space of possible configurations indicating which configuration is collision-free, is compatible with the implicit kinematic model. Given the neural density field and information about obstacle(s) in the scene, a configuration is valid if the maximum density of the robot among $N$ uniformly sampled points from the volume of the obstacle $O_v$ is below a threshold. Thus, sampling-based motion planning methods that 
search the configuration space, %
such as Probabilistic Roadmap \cite{kavraki1996probabilistic} or Rapidly-exploring Random Trees (RRT) \cite{lavalle1998rapidly}, can be used.
The membership in the configuration space is queried as:
\begin{equation}
    Q(\boldsymbol{\theta}) = \begin{cases}
        True & \text{if:} ~ \max_{\mathbf{x}\in O_v} \left [ \alpha(f_\Phi(\mathbf{x},\boldsymbol{\theta})) \right ] < \tau \\
        False & \text{else.}
    \end{cases}
\end{equation}

\section{Experimental Setup}

\begin{figure*}[!htb]
\begin{center}
\vspace{1.5mm}
\includegraphics[scale=0.445]{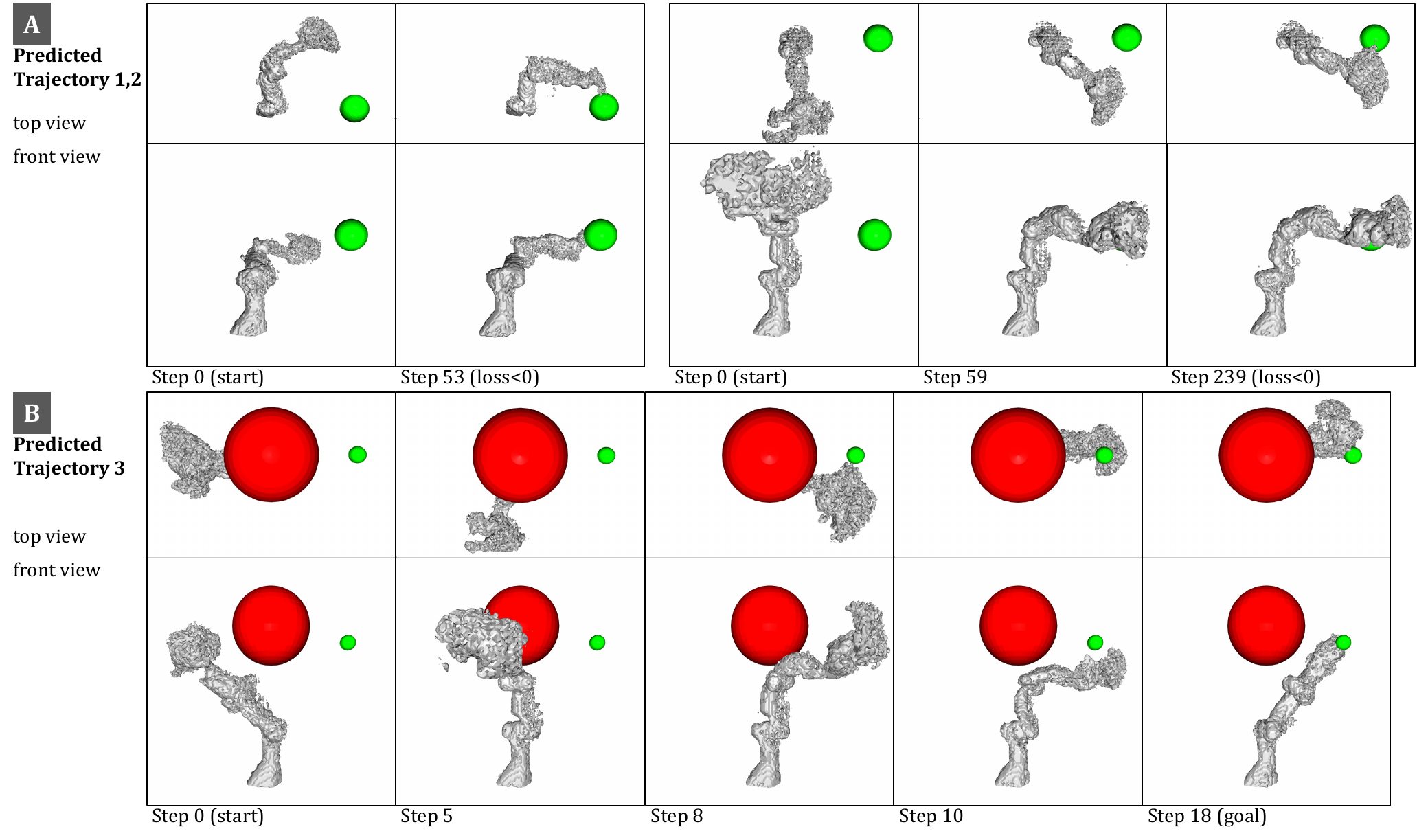}
\end{center}
   \caption{\textbf{Motion planning results:} A) Joint value optimization via input PGD. A density loss is minimized when the sphere (green) is touched. B) RRT planning in config. space to intersect a point (green). The query model rejects samples with density on the obstacle (red). Please also see suppl. video \cite{video}. 
   }
\label{fig:touch}
\end{figure*}

We %
demonstrate our method on a simulated 7-DOF robot.

\textbf{Training distribution.} We apply the curriculum data generation described above with 16 different %
configurations per set $s$ and 6 random base rotations sampled anew for each configuration, totalling $5,588$ annotated $400\times400$ images. %
We generate the images in simulation of the Panda robot \cite{pbrobots} with 7 joints and a rotatable base ($k=8$), using the Pybullet simulator \cite{coumans2016pybullet}. We group batches of 15 images and, to avoid overfitting to one batch, only process $10,240$ rays per image.

\textbf{Training.} We use the described architecture with 3-layer DOF-individual encoder MLPs, 1-layer coordinate encoder MLPs, 2-layer group encoder MLPs, and the final 7-layer density MLP. We train using the Adam optimizer for 1,320,000 steps with a learning rate of $4e-5$ and optimize the parameters of all MLPs together.

\textbf{Visualizations.} 
To visualize the \textit{predicted} self-model, we produce point clouds by querying the field from two camera poses at the front and side of the scene on the $y$- and $x$-axes. Points with alpha values above $0.015$ are kept, determining the isolevel. On the fused point cloud, marching cubes \cite{lewiner2003efficient} reconstruction is applied to generate a triangle mesh, followed by hole repair and Taubin smoothing \cite{taubin1995curve} algorithms.
To visualize the \textit{ground truth}, we simulate the true robot model in Pybullet. The ground-truth point cloud for a joint configuration is the fusion of six point clouds from RGB-D images obtained from two camera views per axis, one at either end, with the view centered at the object. The mesh is then produced identically to the predicted mesh.

\textbf{Metrics.} To assess the quality of our model, we compare the ground-truth against the predicted meshes. First, we use the customary Chamfer-L2 distance, the shortest Euclidean distance of a point in a set to any point in the other set, applied symmetrically and averaged over all points. This returns an average spatial offset per point. We generate the point sets by sampling uniformly from the mesh surfaces. 
Second, as measures for the spatial similarity of the shapes,
we compute two intersection over union (IoU) metrics. Both are based on a union point cloud, constructed by fusion, and an intersection point cloud, constructed by keeping points with a negative signed distance to the mesh defined by the other point cloud. We compute a 2D metric, relating the surface areas of the meshes reconstructed from the point clouds, and a 3D metric, relating the volumes of their convex hulls, %
produced from %
uniformly sampled surface points.

\section{Results}

\textbf{Neural-field self-model.}
We show the predicted meshes from the 7-DOF robot self-model for five random test configurations from a fixed view in Fig. \ref{fig:meshes}. It can be observed that in each configuration, the prediction follows the shape of the ground truth, subject to small deviations in the rotations of smaller parts of the body. For the shown samples, this indicates that the model learned to correctly approximate the shape from the configuration, despite the large space of possible configurations. 
We find that density scales with the certainty in the prediction and that despite the solid material of the robot, most of the non-zero density values are in the lower regime as opposed to close to one. Consequently, the threshold selection, that is the marching cube isolevel, is a significant hyperparameter since it controls the sensitivity with which sampled points are included. %
A too high value may exclude parts of the body in whose prediction the model is less certain. Due to the serial  dependence among the DOFs - the first link's density only depends on the first joint value, whereas the seventh link's density depends on all previous joint values - those excluded parts are the upper parts of the body. The highest density values belong to points in the base of the robot, which remains static. In addition, the querying resolution determines the trade-off between computational cost and approximation of the true predicted model.

In addition to the qualitative evaluation, numerical results on the spatial distance metrics are provided in Table \ref{tab:results}.
As the most important metric, the mean of the Chamfer-L2 distance for the test set is 1.94\% relative to the length of the shortest dimension of the workspace of the robot, 1.254m along the vertical axis. This indicates that, on average, each point on the mesh that was reconstructed from the points in the volume predicted to have sufficiently high density is close to a point on the robot's true surface given the queried configuration. Greater variance can be observed for the two IoU metrics. For the volume-based IoU, this is due to the constraint that the hull must convexly contain all points of the surface point cloud so that outliers have a large effect on its shape and, thus, volume. In addition, while marginally off-positioned predicted robot parts can still produce moderate Chamfer-L2 distances, these parts may not or only partially intersect with the ground-truth parts, leading to a lower value. The surface area is similarly outlier-sensitive and depends on the smoothness of the surface, which is not reliably given.

\textbf{Motion planning.}
In Fig. \ref{fig:touch}, PGD- and RRT-generated trajectories are shown. The task for the former was to touch %
an object in an obstacle-free environment, while the task for the latter was to circumvent an obstacle to move from a start to a goal position. 
For PGD, the joint-limit-projected optimization results in valid trajectory steps. %
The robot moves itself into a configuration in which the sphere is touched such that the density on a surface point is above the threshold ($\tau=0.6$). Unlike in classical kinematics, the part touching the target can be different from the end effector. However, distant start configurations may stop in local optima before reaching the target, rendering the learning rate a crucial hyperparameter. In addition, $\tau$ controls the closeness to the target in the final configuration.
In those challenging cases, neural-field-based RRT reliably finds valid trajectories. In Fig. \ref{fig:touch}, the robot is able to move around the obstacle to reach its goal position. This approach is computationally more expensive since the neural field is queried extensively to construct the tree. Similarly, the strategy for sampling from the obstacle's volume impacts the performance and runtime. %

\section{Conclusion and Outlook}
We propose dynamic neural density fields conditioned on high DOFs. %
To this end, we introduce a hierarchical MLP architecture and curricular data sampling strategy. We use this method to learn the first neural-implicit self-model of a robot without depth or geometry information and from one camera, which can be used in lieu of a classical kinematic model. %
Future work may explore limiting the training data to more sparsely observed DOFs configurations and removing the need for camera parameter annotation via automatic estimation. %
In navigation and manipulation tasks, the integration of our approach, which models the robot, with previous work, which models the robot's environment, would be beneficial, as well as an extension to multi-robot setups.
We highlight the usability of our method for dynamic object-centric scenes outside robotics in general DOFs-controlled environments.

\section{Acknowledgment}
This work was supported in part by the US National Science Foundation (NSF) AI Institute for Dynamical Systems (DynamicsAI.org), grant 2112085.

{\small
\printbibliography
}

\end{document}